\documentclass[sigconf,screen]{acmart}
\usepackage{url}
\usepackage{xcolor}
\definecolor{newcolor}{rgb}{.8,.349,.1}
\usepackage{enumitem}
\usepackage{colortbl}
\usepackage{float}
\usepackage[ruled]{algorithm2e}
\usepackage{subcaption}
\usepackage{multirow}
\usepackage{enumitem}
\definecolor{mygray}{gray}{.9}
\definecolor{ForestGreen}{RGB}{34,139,34}

\usepackage{tcolorbox}
\copyrightyear{2024}
\acmYear{2024}
\setcopyright{rightsretained}
\acmConference[MRAC '24] {Proceedings of the 2nd International Workshop on Multimodal and Responsible Affective Computing}{November 1, 2024}{Melbourne, VIC, Australia.}
\acmBooktitle{Proceedings of the 2nd International Workshop on Multimodal and Responsible Affective Computing (MRAC '24), November 1, 2024, Melbourne, VIC, Australia}
\acmDOI{10.1145/3689092.3690042}
\acmISBN{979-8-4007-1203-6/24/10}

\makeatletter
\gdef\@copyrightpermission{
  \begin{minipage}{0.3\columnwidth}
   \href{https://creativecommons.org/licenses/by/4.0/}{\includegraphics[width=0.90\textwidth]{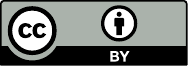}}
  \end{minipage}\hfill
  \begin{minipage}{0.7\columnwidth}
   \href{https://creativecommons.org/licenses/by/4.0/}{This work is licensed under a Creative Commons Attribution International 4.0 License.}
  \end{minipage}
  \vspace{5pt}
}
\makeatother

\begin{document}

\title{MRAC Track 1: 2nd Workshop on Multimodal, Generative and Responsible Affective Computing}

\author{Shreya Ghosh}
\orcid{0000-0002-2639-8374}
\affiliation{%
  \institution{Curtin University}
  \city{Perth}
  \country{Australia}}
\email{shreya.ghosh@curtin.edu.au}

\author{Zhixi Cai}
\orcid{0000-0001-7978-0860}
\affiliation{%
  \institution{Monash University}
  \city{Melbourne}
  \country{Australia}}
\email{zhixi.cai@monash.edu}

\author{Abhinav Dhall}
\orcid{0000-0002-2230-1440}
\affiliation{%
  \institution{Flinders University}
  \city{Adelaide}
  \country{Australia}}
\email{abhinav.dhall@flinders.edu.au}

\author{Dimitrios Kollias}
\orcid{0000-0002-8188-3751}
\affiliation{%
  \institution{Queen Mary University of London}
  \city{London}
  \country{United Kingdom}}
\email{d.kollias@qmul.ac.uk}

\author{Roland Goecke}
\orcid{0000-0003-2279-7041}
\affiliation{%
  \institution{UNSW}
  \city{Canberra}
  \country{Australia}}
\email{r.goecke@unsw.edu.au}

\author{Tom Gedeon}
\orcid{0000-0001-8356-4909}
\affiliation{%
  \institution{Curtin University}
  \city{Perth}
  \country{Australia}}
\email{tom.gedeon@curtin.edu.au}

\renewcommand{\shortauthors}{Shreya Ghosh et al.}
\begin{abstract}
With the rapid advancements in multimodal generative technology, Affective Computing research has provoked discussion about the potential consequences of AI systems equipped with emotional intelligence. Affective Computing involves the design, evaluation, and implementation of Emotion AI and related technologies aimed at improving people's lives. Designing a computational model in affective computing requires vast amounts of multimodal data, including RGB images, video, audio, text, and physiological signals. Moreover, Affective Computing research is deeply engaged with ethical considerations at various stages-from training emotionally intelligent models on large-scale human data to deploying these models in specific applications. Fundamentally, the development of any AI system must prioritize its impact on humans, aiming to augment and enhance human abilities rather than replace them, while drawing inspiration from human intelligence in a safe and responsible manner. The MRAC 2024 Track 1 workshop seeks to extend these principles from controlled, small-scale lab environments to real-world, large-scale contexts, emphasizing responsible development. The workshop also aims to highlight the potential implications of generative technology, along with the ethical consequences of its use, to researchers and industry professionals. To the best of our knowledge, this is the first workshop series to comprehensively address the full spectrum of multimodal, generative affective computing from a responsible AI perspective, and this is the second iteration of this workshop.
Webpage: \url{https://react-ws.github.io/2024/}
\end{abstract}





\maketitle

\section{Introduction}

Affective Computing is a multidisciplinary field that involves the sensing, computational modelling, evaluation, and deployment of AI systems capable of recognizing, interpreting, and responding to human emotions~\cite{picard2000affective}. These systems rely on large-scale multimodal data encompassing RGB images, video, audio, text, and physiological signals to develop accurate and effective emotion-AI models~\cite{ghosh2023emolysis,kollias20247th,kollias2019expression,cai2023marlin}. The goal is not just to create machines that can mimic human emotions, but to develop systems that can genuinely enhance human communication, interactions~\cite{sharma2023graphitti,ghosh2018role,ghosh2020automatic,jahangard2024jrdb} by being more emotionally intelligent.

The computational modelling begins with collecting and annotating multimodal datasets, developing algorithms that can process and interpret this data, and training models to recognize and respond to emotional cues. This phase is heavily reliant on data that captures human non-verbal cues, including facial expressions, eye-gaze, voice intonations, body language, and even physiological signals like heart rate or skin conductivity~\cite{hasan2023empathy,ghosh2023automatic_,dhall2016emotion,li2020deep,sharma2013modeling,zhu2019detecting,yao2018deep}. The challenge here is not just in the technical development but also in ensuring that the data used in the training process is representative, diverse, and ethically sourced~\cite{devillers2023ethical}. After computational modelling, the evaluation phase is critical in ensuring that the Emotion AI models perform as intended across different scenarios and populations. This involves rigorous testing with diverse datasets to ensure that the models can accurately recognize and respond to a wide range of empathetic responses. The evaluation process must also consider the ethical implications of these technologies, such as potential biases in emotion recognition, the impact on users' privacy, and the consequences of incorrect or inappropriate emotional responses by the AI~\cite{jiang2023generalised,jiang2024towards}. Once the models have been created and evaluated, the final stage is deployment, where these systems are integrated into real-world applications~\cite{amin2024wide,lu2024gpt}. This could range from customer service bots that can detect and respond to customer frustration~\cite{gupta2024facial}, to healthcare applications that monitor patients' emotional states to provide better care~\cite{li2021plaan,liu2024affective}. However, deployment brings its own set of ethical challenges, such as ensuring that these systems are used in ways that are beneficial to users and do not infringe on their rights or well-being~\cite{cortinas2023toward,liu2024affective}. Basically, the field of Affective Computing mainly aim to integrate emotional intelligence to computers or AI systems. As the development and deployment of Emotion AI and related technologies progress, they offer the potential to significantly enhance human life by making machines more responsive and empathetic. The MRAC 2024 workshop aims to address these by focusing on the responsible creation, evaluation, and deployment of Emotion AI and assistive technologies with applications in education, entertainment and healthcare.


\subsection{Responsible Affective Computing}
Ethics is the building block of any responsible AI systems, and Affective Computing is no exception. The development of AI systems that can recognize and respond to human emotions must consider ethical aspects. 

\noindent \textbf{Data Privacy and Consent.} The collection and use of multimodal data raise concerns about privacy and consent. For instance, the use of facial recognition technology to detect emotions can be invasive, especially if users are not fully aware of how their data is being used. Ensuring that data is collected with informed consent and users have control over their data is crucial in the responsible development of Emotional AI~\cite{van2023privacy,liu2020disguising}.

\noindent \textbf{Bias and Fairness.} Emotion AI systems learns on the basis of input training data. If the training data is biased whether in terms of demographic representation or cultural context; the resulting computational models can propagate these biases~\cite{yang2022enhancing}. This could lead to scenarios where certain groups are unfairly treated or misinterpreted by the AI, leading to negative consequences~\cite{sham2023ethical}. Addressing these biases is critical to developing fair and inclusive Emotion AI systems for different applications~\cite{sogancioglu2023effects,sogancioglu2023using}. Also, there is an increasing risk of generated data being used `without knowing it is generated' in training models. There is always a potential risk associated with it.

\noindent \textbf{Transparency and Interpretability.} Another key consideration while developing Emotion AI models is the transparency of the system. Users should be aware of how the system work, what data is used, and how decisions are made. This transparency is crucial for building trust in the technology~\cite{sousa2024converging}. 

\subsection{The Role of Generative Technology}
Generative AI, that involves the creation of new data or multimodal digital content by AI systems/models, is becoming increasingly important in Affective Computing~\cite{wang2021building}. For example, generative models can create realistic avatars that can interact with users in a more human-like way. AI can even generate synthetic data to train emotion-AI models. However, the use of generative technology also raises additional concerns.

\noindent \textbf{Synthetic Data and Deepfakes.} The ability to generate synthetic data can be beneficial for training Emotion AI models, especially when real-world data is scarce or difficult to obtain. However, this same technology can also be used to create deepfakes manipulated videos or images that can be used to deceive or harm others. Ensuring that generative technology is used responsibly to prevent potential misuse is a critical challenge~\cite{cai2023av,cai2023glitch,narayan2023df,cai2022you}.

\noindent \textbf{Human-Machine Interaction.} It is a growing field in the area of affective computing that aims to encode human non verbal as well as verbal behavior~\cite{ghosh2019predicting,sharma2019automatic,ghosh2018automatic} followed by incorporating it in human AI interaction systems~\cite{amershi2019guidelines} using generative technologies. 

\subsection{MRAC 2024 Workshop: A Focus on Multimodal, Responsibile and Generative AI}
The MRAC 2024 workshop aims to address these challenges by focusing on responsible, multimodal and generative Affective Computing. The workshop explore how to transfer the principles of responsive and responsible AI from small-scale, lab-based environments to real-world, large-scale applications. This includes not only the technical aspects of creating and deploying Emotion AI systems but also the broader  implications of the human centric technologies.

\noindent \textbf{Real-World Applications.} The workshop showcase real-world applications of Affective Computing in healthcare, education, entertainment etc, highlighting the potential benefits and the challenges involved. We believe that participants have the opportunity to discuss how to design Emotion AI systems that are both effective and ethically sound, with a focus on transparency, explainability, and user empowerment.

\noindent \textbf{Interdisciplinary Collaboration.} The workshop will bring together researchers, industry professionals to discuss the future of Affective Computing. By fostering interdisciplinary collaboration, the workshop aims to ensure that the development of Emotion AI is guided by diverse perspectives and the inclusiveness is integrated into every stage of the process.

\section{Objective, scope and topics of the workshop}
The 2nd International Workshop on Multimodal, Generative and Responsible Affective Computing (MRAC 2024) at ACM-MM 2024 (track for Multimodal and Responsible Affective Computing) aims to encourage and highlight novel strategies for affective phenomena estimation and prediction with a focus on robustness and accuracy in extended parameter spaces, spatially, temporally, spatio-temporally and most importantly `Responsibly'. This is expected to be achieved by applying novel neural network architectures, generative ai, incorporating anatomical insights and constraints, introducing new and challenging datasets, and exploiting multi-modal training. The topics of the workshop include but not limited to the following:
\begin{enumerate}
    \item \textit{Large scale data generation or Inexpensive annotation for Affective Computing}
    \item \textit{Generative AI for Affective Computing using multimodal signals}
    \item \textit{Multi-modal method for emotion recognition}
    \item  \textit{Privacy preserving large scale emotion recognition in the wild}
    \item \textit{Generative aspects of affect analysis}
    \item \textit{Contextual Gesture Generation}
    \item \textit{Deepfake generation, detection and temporal deepfake localization aka Gernerative AI}
    \item \textit{Multimodal data analysis}
    \item \textit{Affective Computing Applications in education, entertainment \& healthcare}
    \item \textit{Explainable or Privacy Preserving AI in affective computing}
    \item \textit{Generative and responsible personalization of affective phenomena}
    \item \textit{Zero-shot and few-shot learning strategies in Emotion AI data}
    \item \textit{Bias in affective computing data (e.g. lack of multi-cultural datasets)}
    \item \textit{Affective Video Understanding}
    \item \textit{Semi- supervised, weakly supervised, unsupervised, self-supervised learning methods}
    \item \textit{domain adaptation methods for Affective Computing }
\end{enumerate}

\section{Research Impact, Relevance and Expert Talks}

\noindent \textbf{Impact.}  In recent years, the AI revolution has already influenced our daily life through virtual assistants across various sectors such as healthcare, banking, transportation, and education. As we look toward the future, it's clear that humans will increasingly interact with AI-powered systems, potentially even more frequently than with other humans. 

To this end, Affective computing holds numerous promising robust future applications, such as forecasting and preventing anxiety, stress, depression, and other mental health issues~\cite{ghosh2021depression}; enhancing robotic empathy~\cite{hasan2023empathy}; supporting individuals with communication, behavioral, and emotional regulation challenges; and promoting overall health and well-being. However, these applications often involve handling sensitive, private, and personal data, necessitating strong controls and protections. Therefore, it is essential to consider emotionally intelligent systems that are `Responsive' and `Responsible'.

\noindent \textbf{Relevance.} Affective Computing often involves unimodal or multimodal data such as images, video, audio, text, physiological signals. Moreover, for the adaptation process to a new deployment scenario requires environment, culture specific data. Developing AI models with multimodal human data is highly co-related with the themes of ACM-MM 2024. The five major themes of ACM-MM 2024 are Engaging Users with Multimedia, Experience, Multimedia systems, Multimedia Content Understanding and Multimedia in the Generative AI Era. All of the themes are co-related with this workshop explicitly or implicitly. Engaging users with multimedia aims to encode emotional and social signal which is highly aligned with the workshop theme. Similar aims with understanding multimedia context (where the ground truth is human-level perception), experience (interaction and experience), generative AI for affective face generation and multimedia system.

\noindent \textbf{Expert Talks.} We invited Prof. Julian Epps (University of New South Wales)~\cite{epps2024wearable} and Prof. Mohammed Bennamoun (University of Western Australia)~\cite{bennamoun2024seeing} to deliver keynotes. Their keynote details are mentioned in our website.

\section{Contribution papers}
This is the second iteration of MRAC workshop, where the accepted papers mainly focus on potential threats of generative AI and deepfakes~\cite{luca2024wdl,wu2024thefd}, long video based affect understanding~\cite{bai2024can} and affect prediction in video conversation~\cite{jia2024are}. In the remainder of this section, we briefly introduce each accepted paper.

\subsection{Are You Paying Attention? Multimodal Linear Attention Transformers for Affect Prediction in Video Conversations}

The post COVID-19 era has witnessed a significant increase in the adoption of video-based communication, highlighting the importance of unintrusive affect recognition method during digital interactions. This paper presents a multimodal emotion recognition in video conversational settings by leveraging linear attention-based Transformer framework from audio-visual cues. 

This paper~\cite{jia2024are} involves exploring various linear attention mechanisms and comparing them with traditional self-attention techniques. By utilizing the K-EmoCon dataset~\cite{park2020k}, the method demonstrates competitive performance in predicting affective states while significantly enhancing memory efficiency. Ablation studies reveal that carefully optimized simple fusion methods can match or even surpass the effectiveness of more complex approaches. The research contributes to the development of more accessible and efficient multimodal emotion recognition systems tailored for video-based conversations. 

These advancements have practical applications in enhancing remote communication and monitoring digital well-being, particularly in the context of the post-pandemic era where digital interactions have become increasingly prevalent. The findings emphasize the potential for linear attention-based models to offer robust solutions for emotion recognition, providing a balance between performance and computational efficiency. This work underscores the importance of efficient and accessible tools for emotion recognition in the evolving landscape of digital communication. By addressing the challenges of memory efficiency and model complexity, the proposed approach offers a viable pathway for integrating affect recognition into everyday video interactions, thereby contributing to improved remote communication experiences and better digital well-being monitoring.

\subsection{Can Expression Sensitivity Improve Macro- and Micro-Expression Spotting in Long Videos?}

Spotting facial expressions is crucial as it directly reflects emotions, particularly those underlying feelings and intentions that might not be verbally expressed. Detecting both macro- and micro-expressions is especially important in psychological analysis, as it provides insight into nuanced emotional states. The detection process involves identifying specific intervals within a long video that contain either macro- or micro-expressions, distinguishing these critical moments from other facial movements occurring at different time scales. Building on successful outcomes in micro-expression recognition, this paper proposes an algorithm designed to enhance the performance of spotting both macro- and micro-expressions using an expression-sensitive model. 

This study~\cite{bai2024can} introduces a novel and effective approach that emphasizes the integration of recognition and detection tasks in the analysis of facial expressions. By utilizing datasets initially intended for micro-expression recognition, this approach broadens the scope of these datasets, expanding their utility beyond their original purpose. The proposed algorithm leverages a pre-trained CNN model and introduces a unique confidence value-based mechanism to label training samples. This method not only achieves superior performance compared to other advanced techniques but also demonstrates an exceptional balance across various databases, encompassing both macro- and micro-expressions, as well as between precision and recall. This balanced performance is crucial for accurately detecting and analyzing expressions across different contexts. 

The benchmarks and evaluation standards used in this study adhere to the challenge policies and evaluation metrics established by MEGC 2020~\cite{jingting2020megc2020} and MEGC 2021~\cite{li2022megc2022}. By aligning with these standards, the proposed approach ensures that its results are both reliable and comparable with other leading methods in the field. Overall, this study contributes to the advancement of facial expression analysis by offering a robust solution for the detection of both macro- and micro-expressions, enhancing the understanding of subtle emotional cues in psychological and behavioral analysis.

\subsection{THE-FD: Task Hierarchical Emotion-aware for Fake Detection}

The rapid advancement of deepfake generation technology has led to the emergence of more naturally fine-grained modifications within short segments of content. Despite this, most research has primarily focused on classifying entire videos as either real or fake, with only a limited number of studies addressing the precise localization of forgeries within both audio and visual modalities of a video. This paper introduces a comprehensive solution that not only addresses whole-video fake classification but also emphasizes segment-level forgery temporal localization. 

The proposed approach~\cite{wu2024thefd}, Task Hierarchical Emotion-aware for Fake Detection (THE-FD) architecture, is designed to initially process video-level data and seamlessly adapt to segment-level analysis. This hierarchical structure provides a general framework that allows for the inheritance and sharing of features across different levels, optimizing the detection process. Key innovations of the THE-FD architecture include the introduction of an emotional feature space to represent sentiment perturbations, which are often indicative of manipulations. Additionally, a fake-frame heatmap module is proposed to capture hidden fake patterns across multimodal data, enhancing the system's ability to detect subtle alterations. To further refine the detection process, a multiscale pyramid transformer is employed to learn features at various levels and time scales, ensuring comprehensive coverage of potential forgeries. The combination of these advancements results in a significant improvement in performance compared to existing methods. 

The proposed approach not only enhances the accuracy of fake detection but also ranks among the top-performing methods in the 2024 1M-Deepfakes Detection Challenge~\cite{cai_1mdeepfakes}. By addressing the challenges of fine-grained temporal localization and leveraging the emotional feature space, this research contributes to the development of more robust and precise tools for detecting deepfake content, offering a valuable resource in the ongoing battle against digital manipulation.

\subsection{W-TDL: Window-Based Temporal Deepfake Localization}

Nowadays distinguishing highly realistic synthetic data from genuine samples has become increasingly challenging. This growing sophistication is particularly concerning with the rise of deepfake content, including images, videos, and audio, which are often used for malicious purposes, making effective detection methods more critical than ever. Despite significant advancements through various competitions, a substantial gap remains in accurately identifying the temporal boundaries of these manipulations. 

To address this challenge, a novel approach for temporal deepfake localization (TDL)~\cite{luca2024wdl} is proposed, utilizing a window-based method for audio (W-TDL) alongside a complementary visual frame-based model. The contributions of this approach are twofold: firstly,~\cite{luca2024wdl} introduces an effective method for detecting and localizing manipulated segments in both video and audio; secondly, it tackles the issue of unbalanced training labels within spoofed audio datasets, which often hinder the performance of detection models. Leveraging the EVA visual transformer for precise frame-level analysis enhances the ability to identify deepfake manipulations in video content. In addition, the modified TDL method for audio improves the detection of temporal boundaries, ensuring more accurate identification of fake segments. This dual-model strategy addresses current limitations in deepfake detection, offering a comprehensive solution that integrates both visual and auditory cues. 

Comprehensive experiments on the AV-Deepfake1M dataset have demonstrated the effectiveness of this approach, achieving competitive results in the 1M-DeepFakes Detection Challenge~\cite{cai_1mdeepfakes}. The success of this method underscores its potential as a robust solution for detecting and localizing deepfake manipulations, marking a significant advancement in the fight against deepfake threats. This work aims to contribute to the development of more secure and reliable systems for identifying synthetic content across various media types.

\section{Summary}
The MRAC 2024 workshop aims to pave the way for the responsible use of multimodal and generative technologies in Affective Computing. Affective Computing represents a powerful tool for enhancing human interactions with AI systems by adding the ``empathy'' component. As multimodal and generative technologies continue to advance, it is crucial to ensure that these systems are developed responsibly, with a focus on augmenting human capabilities. The development of Emotion AI relies on vast amounts of multimodal data, including images, video, audio, text, and physiological signals, to create models capable of recognizing and interpreting human emotions. While this data is crucial for the accuracy and effectiveness of these systems, it also raises concerns about privacy, consent, and potential biases. To ensure that these models are trained on diverse, representative datasets is essential to avoid propagating biases and to create fair, inclusive AI systems.

\section{Future Directions}
The future of multimodal, generative and responsible AI holds several exciting possibilities and challenges as affective computing and assistive technology continues to advance. Here are some key aspects to consider while thinking about the future of emotion AI:

\begin{enumerate}
    \item \textbf{Multi-modal/Cross-modal Analysis:} Over the past decade, head, eye and facial gesture synthesis has become an interesting line of research. Prior works in this area have mainly use visual modality to generate realistic gesture. The main challenge in this domain is multimodal fine grained data annotation for motion synthesis, gesture analysis which is a noisy and error prone process. Research along this direction have potential to estimate assistive technology in challenging situation where visual stimuli fails.

    \item \textbf{Emotion AI in Unconstrained Settings:} The most precise methods for emotion and gesture estimation is performed via intrusive sensors, IR camera, and RGBD camera. The main drawback of existing systems is that their performance degrades when used in real-world settings. In future, assistive technology models should consider these situations. Although several current efforts in this direction employ techniques, yet further research is needed. Moreover, most of the current affective computing benchmark datasets require the proper geometric arrangement as well as user cooperation. It would be an interesting direction to explore it in a more flexible setting.
    
    \item \textbf{Learning with Limited Supervision and LLMs:} With the surge in unsupervised, self-supervised, weakly supervised and VLM techniques in this domain, more exploration in this direction is required to reduce or eliminate the dependency on ground truth labels which could be error-prone due to data acquisition limitations.
    
    \item \textbf{Bias and Fairness Mitigation:} Efforts to mitigate bias in AI will continue to evolve. Advanced techniques, such as adversarial debiasing and fairness-aware machine learning, will be developed to reduce bias in AI systems and ensure fairness in models's decision-making processes.

    \item \textbf{Explainability and Transparency:} The demand for AI explainability and transparency will intensify. Researchers will work on making AI models more interpretable and understandable, helping users to trust AI systems and better understand model's decision-making processes.

    \item \textbf{Human-AI Collaboration:} The future of responsive AI will involve closer collaboration between humans and AI systems. Human-AI partnerships will become more integrated into various domains, enhancing productivity, decision-making, and problem-solving capabilities.

    \item \textbf{Privacy-Preserving AI:} Techniques for privacy-preserving AI will continue to advance, allowing AI to work with sensitive data while protecting individuals' privacy rights. Federated learning, differential privacy, and secure multiparty computation are examples of privacy-preserving approaches. A new approach is the generation of synthetic data which matches important distributional properties of private data, and can be used to train models which are then tested on the private data which therefore can solve the purpose of keeping the real data truly private. 

    \item \textbf{Ethical AI Research:} There will be a growing emphasis on conducting research specifically focused on the ethical aspects of AI, including the exploration of novel ethical theories, frameworks, and guidelines tailored to AI systems. Companies and organizations will be expected to demonstrate a strong commitment to responsible AI practices. Ethical AI development and deployment will become a competitive advantage, and consumers may prioritize products and services that align with their values. As AI intersects with emerging technologies like quantum computing, biotechnology, and autonomous vehicles, ethical considerations will become even more critical. Responsible AI frameworks and guardrails will need to adapt to these evolving technologies.

    \item \textbf{Generative AI addressing inclusion:} A major challenge in generative AI for human centric computing comes from the variability due to different cultures and languages. By focusing on universal features and building language-agnostic systems, generative AI can become more effective in empowering ML models, even in scenarios where the language or cultural context is unfamiliar. From a generative AI perspective, generating audio-visual content in non-native languages presents a unique challenge. Typically, when foreign language videos are accessed, they are often accompanied by voice-overs or native language subtitles. This creates a complex scenario where the subtleties of the original language, cultural information, and context may be missed, making it difficult for both humans and AI models to accurately interpret the video. Thus, it is crucial to develop and use diverse datasets containing generative content in multiple languages. Additionally, the models should be trained to prioritize features that are invariant to language, such as discrepancies in audio-visual synchronization or unnatural facial movements, allowing them to generalize across different linguistic and cultural scenarios.

\end{enumerate}

The future of emotion AI holds the promise of more ethical, transparent, and accountable AI systems. However, it also poses challenges related to the evolving nature of technology, the need for adaptive ethical frameworks, and the importance of global collaboration. Emotion AI will continue to be a dynamic field, evolving to meet the ethical and societal demands of AI in an ever-changing world.



\bibliographystyle{ACM-Reference-Format}
\bibliography{main}

\end{document}